# A 3D Multi-Style Cross-Modality Segmentation Framework for Segmenting Vestibular Schwannoma and Cochlea

Yuzhou Zhuang[1](✉)

[1] School of Computer Science and Technology, Huazhong University of Science and Technology, Wuhan, China
zhuang_yuzhou@hust.edu.cn

**Abstract.** The crossMoDA2023 challenge aims to segment the vestibular schwannoma (sub-divided into intra- and extra-meatal components) and cochlea regions of unlabeled hrT2 scans by leveraging labeled ceT1 scans. In this work, we proposed a 3D multi-style cross-modality segmentation framework for the crossMoDA2023 challenge, including the multi-style translation and self-training segmentation phases. Considering heterogeneous distributions and various image sizes in multi-institutional scans, we first utilize the min-max normalization, voxel size resampling, and center cropping to obtain fixed-size sub-volumes from ceT1 and hrT2 scans for training. Then, we perform the multi-style image translation phase to overcome the intensity distribution discrepancy between unpaired multi-modal scans. Specifically, we design three different translation networks with 2D or 2.5D inputs to generate multi-style and realistic target-like volumes from labeled ceT1 volumes. Finally, we perform the self-training volumetric segmentation phase in the target domain, which employs the nnU-Net framework and iterative self-training method using pseudo-labels for training accurate segmentation models in the unlabeled target domain. On the crossMoDA2023 validation dataset, our method produces promising results and achieves the mean DSC values of 72.78% and 80.64% and ASSD values of 5.85 mm and 0.25 mm for VS tumor and cochlea regions, respectively. Moreover, for intra- and extra-meatal regions, our method achieves the DSC values of 59.77% and 77.14%, respectively.

## 1      Introduction

The CrossMoDA challenge [1] provides the first large-scale unpaired 3D cross-modality segmentation dataset, which aims to train segmentation models from labeled contrast-enhanced T1 (ceT1) scans to segment the vestibular schwannoma (VS) and cochlea regions of unlabeled high-resolution T2 (hrT2). The CrossMoDA2023 challenge [2], [3] extends the segmentation task by including multi-institutional, heterogenous data acquired for routine surveillance purposes and introduces a tumor sub-segmentation task of intra- and extra-meatal components, thus generating three segmentation regions (i.e., intra-meatal, extra-meatal, cochlea regions). Unsupervised domain adaptation (UDA) [4] can transfer the effective knowledge learned from the labeled source domain to the unlabeled target domain in an unsupervised manner, which significantly alleviates the domain shift problem between labeled ceT1 and unlabeled hrT2 scans. The Cycle-consistent Generative Adversarial Network (CycleGAN) [5] with cycle-consistency constraint and Contrastive Unpaired Image Translation Network (CUT) [6] with contrastive loss are the two most commonly used unpaired image translation networks in existing UDA methods. However, vanilla CycleGAN and CUT lack the volumetric spatial information and preservation of segmentation regions during image translation [7]. Furthermore, the vanilla contrastive loss of CUT repels all negative samples indiscriminately [8], which is apparently sub-optimal as negative samples usually have different similarities with the anchor. In this work, we proposed a 3D multi-style cross-modality segmentation framework for the crossMoDA2023 challenge, including the multi-style translation and self-training segmentation phases. To overcome heterogeneous distributions in multi-institutional scans, we first perform the multi-style image translation phase to generate multi-style and realistic target-like volumes from labeled ceT1 volumes. In the multi-style image translation, we add the auxiliary segmentation loss and segmentation decoder to translation networks for enhancing the translation performance of segmentation regions. Meanwhile, we design a 2D translation network with weighted contrastive loss and 2.5D translation networks with cycle-consistency and vanilla contrastive loss for image translation. Then, we perform the self-training volumetric segmentation phase in the labeled. In the segmentation phase, we employ the nnU-Net framework [9] and iterative self-training method using pseudo-label learning for training robust and accurate 3D segmentation models in the unlabeled target domain. Finally, we employ the sliding window and model ensemble strategy for predicting unlabeled hrT2 scans by trained segmentation models. On the crossMoDA2023 validation dataset, our method gains promising results and achieves the mean DSC values of 72.78% and 80.64% and ASSD values of 5.85 mm and 0.25 mm for VS tumor and cochlea regions, respectively. For intra- and extra-meatal components, our method achieves the DSC values of 59.77% and 77.14%, respectively.

## 2   Proposed Methods

### 2.1   Overall Framework

As shown in Fig. 1, we propose a two-stage cross-modality segmentation framework based on the 'translation-then-segmentation' strategy for segmenting VS and cochlea regions in the CrossMoDA2023 challenge. Our method first generates realistic target-like volumes from labeled source scans by image translation networks, and then leverages labeled target-like volumes to train supervised 3D segmentation networks for segmenting unlabeled target scans. In our method, auxiliary segmentation tasks of regions of interest and multi-style image translation strategies significantly boost the image translation process. Meanwhile, iterative self-training based on pseudo-label learning effectively improves the generalization performance of 3D segmentation models in the target domain.

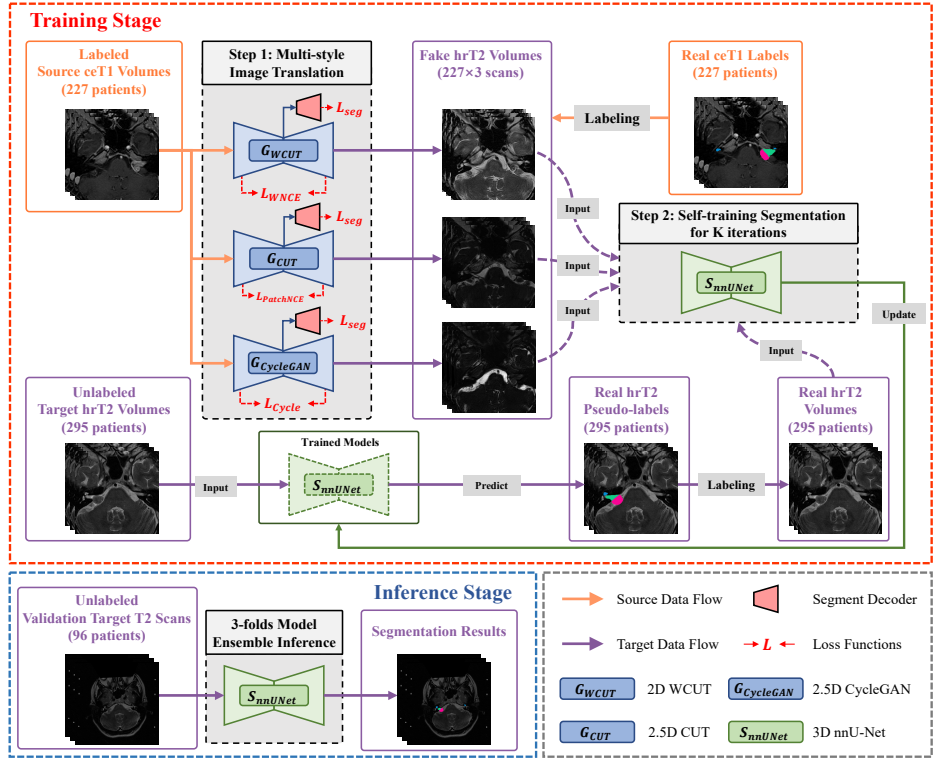

**Fig. 1.** Overall framework of our proposed method, where 'K' is the maximum number of self-training iterations and it is set to 3 by default.

### 2.2 Multi-style Translation

In the multi-style translation phase, we first exploit the three-channel inputs for training 2.5D CycleGAN with the cycle-consistency loss $L_{Cycle}$ and CUT with the contrastive loss $L_{PatchNCE}$, and we add the auxiliary segmentation loss $L_{seg}$ to these models for improving the translation performance of segmentation regions. Then, we design a 2D weighted contrastive unpaired image translation network (WCUT) for generating high-quality target-like volumes. The vanilla PatchNCE loss $L_{PatchNCE}$ of CUT aims to maximize the mutual information between patches in the same spatial location from the synthetic image $X$ and the original image $Y$, which is written as:

$$L_{PatchNCE}(X,Y) = -\sum_{i=1}^{N} \log \frac{\exp(x_i \cdot y_i/\tau)}{\exp(x_i \cdot y_i/\tau) + \sum_{\substack{j=1\\j\neq i}}^{N} \exp(x_i \cdot y_j/\tau)}, \quad (1)$$

where $X=[x_1,\cdots,x_N]$ and $Y=[y_1,\cdots,y_N]$ are encoded image feature sets, $\tau = 0.07$ is the default temperature parameter, $N$ is the number of feature patches. To adjust the pushing force of a negative sample, a simple yet feasible approach is to adjust its weight in the contrastive objective. According to Eq. (1), a higher weight of a negative pair (e.g., $\exp(x_i \cdot y_j/\tau)$) indicates a higher importance in the contrastive objective, i.e., the enlarged pushing force for this negative pair. Thus, we use the WeightNCE loss $L_{WNCE}$ as the contrastive loss of WCUT for a better translation, which can be formulated as:

$$L_{WNCE}(X,Y) = -\sum_{i=1}^{N} \log \frac{\exp(x_i \cdot y_i/\tau)}{\exp(x_i \cdot y_i/\tau) + \sum_{\substack{j=1\\j\neq i}}^{N} w_{ij} \cdot \exp(x_i \cdot y_j/\tau)}, \quad (2)$$

where $w_{ij}$ denotes the weight between sample $y_j$ and anchor $x_i$ and is subjected to $\sum_{\substack{j=1\\j\neq i}}^{N} w_{ij} = 1$, $i \in [1,N]$. The hard weighting weights $w_{ij}$ are determined with a positive and negative relation to the similarity between sample $y_j$ and anchor $x_i$ as below:

$$w_{ij} = \frac{\exp(x_i \cdot y_j/\beta)}{\sum_{j=1}^{N} x_i \cdot y_j/\beta}, \quad (3)$$

where $\beta = 0.1$ denotes the weighting temperature parameter. Finally, we use different trained translation models to predict each slice of the labeled ceT1 volumes on the axial plane continuously, thus generating the labeled fake hrT2 volumes for 3D supervised segmentation.

### 2.3 Self-training Segmentation

In the self-training segmentation phase, we leverage the nnU-Net framework and the iterative self-training strategy to train 3D segmentation models from labeled fake hrT2 volumes and unlabeled real hrT2 volumes, which reduces the distribution gap between real hrT2 and synthetic hrT2 images and to improve the robustness of the segmentation model for unseen real hrT2 scans. The self-training segmentation procedure [10] consists of four steps: (1) Training the segmentation model using the fake hrT2 volumes with labels of real ceT1 volumes; (2) Generating pseudo labels of unlabeled real hrT2 volumes by using the trained segmentation model. (3) Retraining the segmentation model using both the fake hrT2 volumes with labels of the ceT1 volumes and the real hrT2 volumes with pseudo labels. 4) Repeating Steps 2-3 to achieve further performance improvement.

## 2.4 Dataset and Implementation Details

**Dataset.** The CrossMoDA2021 dataset provides 227 labeled ceT1 scans and 295 unlabeled hrT2 scans for training, 96 unlabeled hrT2 scans for online validation, and 365 unlabeled hrT2 scans for testing. The ceT1 and hrT2 scans are multi-institutional, heterogeneous scans from the UK and Tilburg centers, NL. Due to different data sources and imaging parameters, this dataset has heterogeneous distributions and various image sizes, and it has an intra-slice spacing of (0.19~0.86) mm×(0.19~2.2) mm and inter-slice spacing of (0.29~3.48) mm.

**Data preprocess.** For data preprocessing, we first resampled and reoriented all scans to obtain scans with the same orientation of 'LPS' and the voxel size of 1.5mm×0.41mm×0.41mm, and we scaled the intensities of all scans to [-1, 1] by the min-max normalization and range scaling. Then, by computing the central cropping region with intensity higher than the 75-th percentile of the whole volume on the axial plane, we padded and center-cropped each scan to the sub-volume with the size of N×256×256, where N is the number of slices of each scan. Finally, we adopt all sub-volumes with the size of N×256×256 for training translation and segmentation models.

**Implementation details.** Our proposed methods were implemented using the Pytorch framework, and we performed all experiments on an Ubuntu 18.04 workstation with two 24G NVIDIA GeForce RTX 3090 GPUs and an Intel Xeon Gold 5117 2.00 GHz CPU. In the image translation stage, we adopt the encoding and decoding parts of ResNet-based generator with 9 residual blocks in CycleGAN as the translation encoders and decoders in our method, respectively. Meanwhile, the decoder of U-Net is used as the segmentation decoder. During the training process of translation networks, we train the 2D network by single-channel slices with the size of 1×256×256 and train 2.5D networks by three-channel adjacent slices with the size of 3×256×256. We adopt the Adam optimizer and the batch size of 1 to train models for 400 epochs. The initial learning rate is initially set to 0.0002 and linearly decays to 0 during the last 200 epochs. In the segmentation stage, we utilize the nnUNet framework to train our 3D segmentation models by 5-fold cross-validation, and we use the self-training method to improve the segmentation performance in the unlabeled target domain. Specifically, we employ the SGD optimizer with an initial learning rate of 0.01 and a momentum of 0.99 to train segmentation models for 300 Epochs, and the learning rate is gradually reduced by the polynomial learning rate policy. Meanwhile, the batch size is set as 1, and the max number of self-training iterations is set as 3. In the inference stage, we apply the sliding window strategy with an overlap rate of 0.5 and the 3-fold model ensemble strategy to continually predict segmentation probability maps of unlabeled hrT2 scans by the trained segmentation models.

## 3 Results

In the CrossMoDA2023 challenge, the Dice Similarity Coefficiency (DSC) and the Average Symmetric Surface Distances (ASSD) are used to measure the region overlap and boundary distance between the segmentation results and the ground truths, respectively. In our experiments, Table 1 shows the quantitative results of different methods on the crossMoDA2023 validation leaderboard, where 'Multi-style' denotes the multi-style translation strategy using 2D WCUT, 2.5D CycleGAN, and CUT, and 'ST' denotes the self-training method with 5 iterations. From Table 1, by employing 2.5D CycleGAN to generate realistic fake hrT2 volumes for training nnU-Net models, Method #1 obtains the DSC values of 67.70% and 77.64% and ASSD values of 3.42 and 0.53 mm for VS and cochlea regions, respectively. Considering heterogeneous distributions in multi-institutional scans, Method #4 leverages the multi-style translation strategy to alleviate the domain shift between modalities, which significantly improves the DSC values of all regions compared to Method #1. After five self-training iterations, our proposed method (Method #5) gains the best overall DSC value in our experiments, which achieves the DSC values of 72.78% and 80.64% and ASSD values of 5.85 and 0.25 mm for VS tumor and cochlea on the validation dataset, respectively. Furthermore, Method #5 achieves the DSC values of 59.77% and 77.14% for intra- and extra-meatal regions, respectively.

**Table 1.** Quantitative results of different methods on crossMoDA2023 validation leaderboard.

| # | Methods | DSC (%) ↑ | | | | ASSD (mm) ↓ | |
|---|---|---|---|---|---|---|---|
| | | Intra-meatal | Extra-meatal | VS | Cochlea | VS | Cochlea |
| 1 | 2.5D CycleGAN+nnU-Net | 56.00±28.20 | 72.64±26.23 | 67.70±31.59 | 77.64±3.23 | 11.25±38.5 | 0.29±0.13 |
| 2 | 2.5D CycleGAN+CUT+nnU-Net | 55.44±26.81 | 74.24±24.19 | 69.65±29.21 | 79.52±3.38 | 16.19±61.57 | 0.27±0.14 |
| 3 | 2D WCUT+nnU-Net | 57.51±25.41 | 75.44±21.20 | 70.61±27.74 | 78.80±4.38 | **5.52±14.34** | 0.28±0.12 |
| 4 | Multi-style+nnU-Net | 56.67±26.98 | 74.75±23.56 | 70.21±29.12 | 80.37±3.22 | 12.79±15.33 | 0.26±0.13 |
| 5 | Multi-style+nnU-Net+ST (Ours) | **59.77±25.13** | **77.14±23.28** | **72.78±27.56** | **80.64±3.24** | 5.85±14.40 | **0.25±0.13** |

## 4 Conclusions

In this work, we proposed a 3D cross-modality segmentation framework for the CrossMoDA2023 challenge. This framework consists of multi-style translation and self-training segmentation phases. In the translation phase, we exploit three different translation networks with various loss functions and input dimensions for overcoming the intensity distribution discrepancy in unpaired ceT1 and hrT2 scans. In the segmentation phase, we utilize iterative self-training to improve the segmentation performance of 3D segmentation models in the unlabeled hrT2 scans. Experimental results show that our method achieves promising results in the CrossMoDA2023 challenge.